\documentclass[letterpaper,twocolumn,10pt]{article}
\usepackage{usenix2019_v3}

\usepackage{tikz}
\usepackage{amsmath}
\usepackage{url} 

\begin{document}

\date{}

\title{\Large \bf MPP: Model Performance Predictor}


\author{
{\rm Sindhu Ghanta}\\
ParallelM
\and 
{\rm Sriram Subramanian}\\
ParallelM
\and
{\rm Lior Khermosh}\\
ParallelM
\and
{\rm Harshil Shah}\\
ParallelM
\and
{\rm Yakov Goldberg}\\
ParallelM
\and
{\rm Swaminathan Sundararaman}\\
ParallelM
\and
{\rm Drew Roselli}\\
ParallelM
\and
{\rm Nisha Talagala}\\
ParallelM
} 

\maketitle

\begin{abstract}
Operations is a key challenge in the domain of machine learning pipeline deployments involving monitoring and management of real-time prediction quality. Typically, metrics like accuracy, RMSE etc., are used to track the performance of models in deployment. However, these metrics cannot be calculated in production due to the absence of labels. We propose using an ML algorithm, {\em Model Performance Predictor} (MPP), to track the performance of the models in deployment. We argue that an ensemble of such metrics can be used to create a score representing the prediction quality in production. This in turn facilitates formulation and customization of ML alerts, that can be escalated by an operations team to the data science team.
Such a score automates monitoring and enables ML deployments at scale.
\end{abstract}

\section{Introduction}

Using machine learning models to extract insights from massive datasets is a widespread industry goal \cite{IOT-wallstreet}. The training phase typically generates several models and the model with the best predictive performance is deployed to production. However, a model's performance in production depends on both the particular data it receives and the datasets originally used to train the model. Models perform optimally on different data distributions and vary in their capacities for generalization. Production datasets often vary with external factors \cite{modelSelection,driftEnsemble}. Whether rapid or gradual, these variations can require models to be updated or rolled back to maintain good predictive performance. Massive scale in production systems prohibits manual intervention or monitoring of such events, requiring in turn automated methods to detect, diagnose, and improve the quality of predictive performance. However, typical production scenarios do not have real-time labels, so popular metrics that compare predictions with labels cannot be used to assess real-time health. 

We present a technique to track the predictive performance of the deployed models called: {\em Model Performance Predictor} (MPP). It tracks the predictive performance metric of the model. For (a) classification and (b) regression, we present an example that targets (a) accuracy and (b) RMSE respectively as the metric to track. 

Detecting the applicability of an activity model to a different domain
using another model was proposed in \cite{chem} using algorithm-specific information. Similar to our approach, an error dataset is used to train another model, but it is limited to a specific algorithm (random forest) and a unique domain. With a similar goal of detecting the confidence in predictions made by a machine learning algorithm, \cite{conformal} proposed hedging the predictions using conformal predictors. A hold out set (in contrast to error set) is used to obtain a bound on the error probability. 
On the other hand, we present an approach that models the errors by using them as the labels. On similar lines, \cite{health} presented a metric that tracks the divergence in data patterns between training and inference. We argue that an ensemble of such approaches can be customized to serve as a score, based on which alerts can be raised. 

\section{Model Performance Predictor}

The goal of Model Performance Predictor (MPP) algorithm is to predict the predictive performance of the deployed algorithm on the test data. This algorithm is trained on the error dataset which consists of prediction errors made by the primary algorithm. In the training phase, the data is divided into training and validation datasets (apart from the test set). The training dataset is used to train the primary algorithm that will be deployed in production. Predictions made by this algorithm on the validation dataset generate the errors that are used as labels to train the MPP algorithm.

Figure \ref{fig:secon} describes the structure of this framework. Labels of this error dataset are the errors in predictions made by the primary algorithm, and features could be a range of things depending on the application. They could simply be the primary algorithm features themselves, predictions from the primary algorithm, probability measures from the primary predictions or some algorithm-specific metrics, such as the number of trees or variation in output from different trees in a Random Forest. Both primary and MPP algorithms make predictions on the test dataset. The primary algorithm focuses on the classification/regression task, while the MPP algorithm focuses on predicting the performance of the primary algorithm. We present MPP as a binary classification algorithm that predicts whether a prediction is correct (1) or incorrect (0).

\begin{figure}[htp!]
\centering
\includegraphics[trim={0cm 0.1cm 0cm 0cm},clip,width=0.4\textwidth]{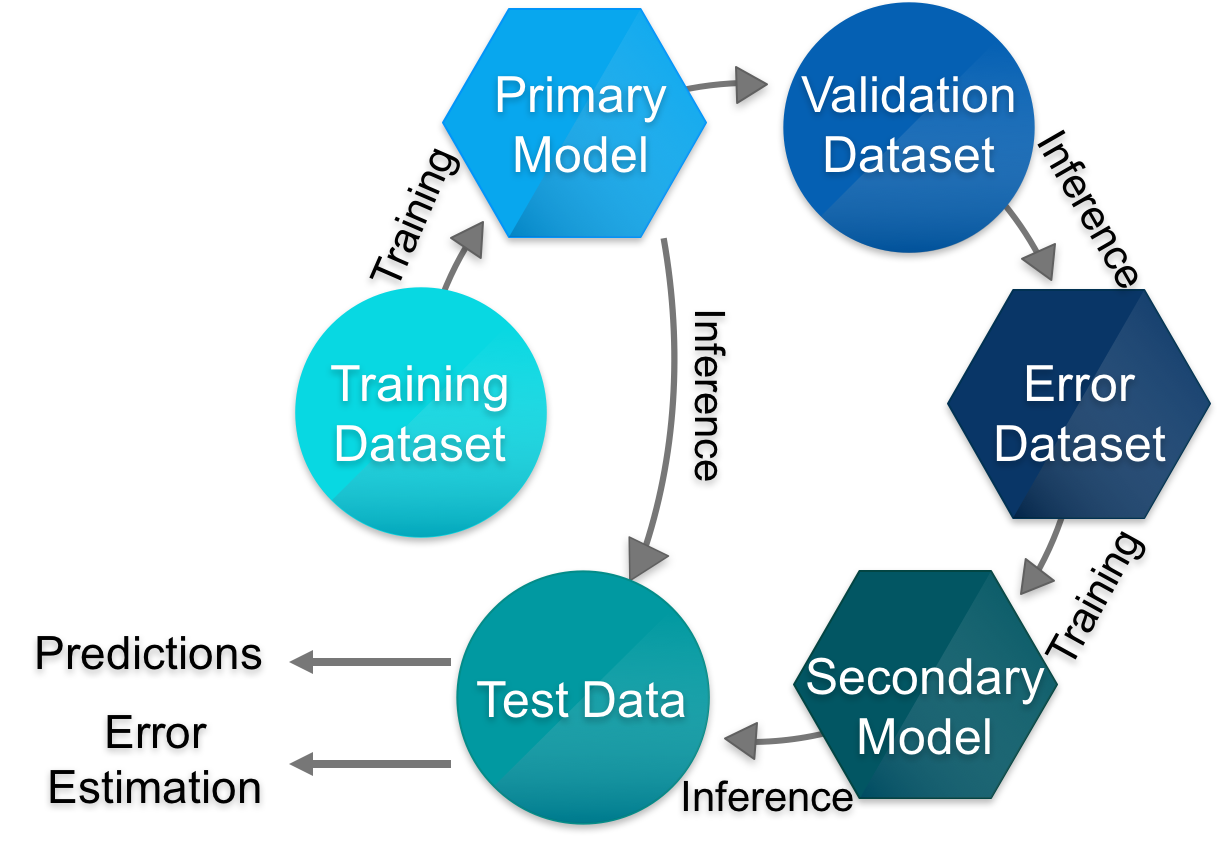}           
\caption{MPP algorithm flow \label{fig:secon}}
\label{fig:secon}
\vspace{-0.5cm}
\end{figure}

\begin{figure}[htp!]
\centering
\includegraphics[trim={0cm 0.1cm 0cm 0cm},clip,width=0.4\textwidth]{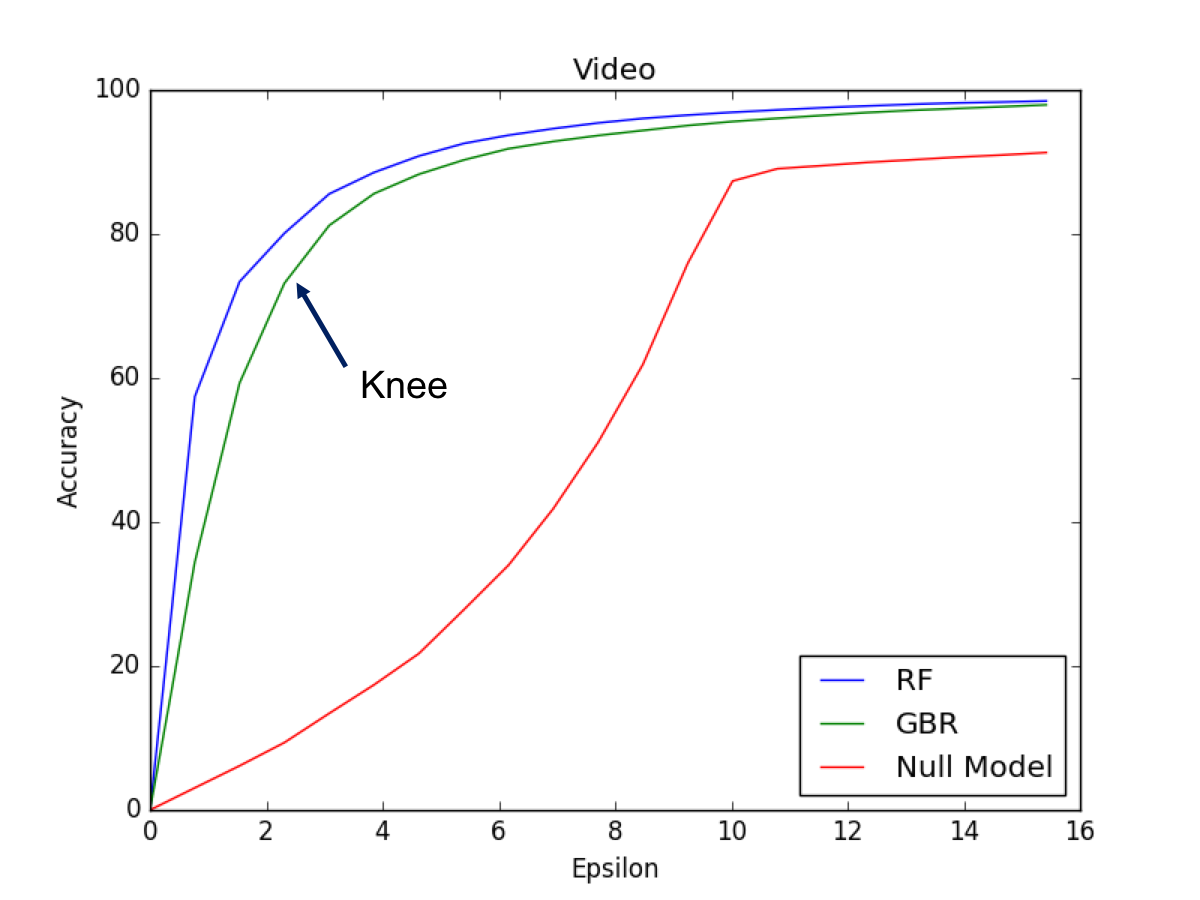}   
\caption{REC curve for the Video dataset \label{fig:rec}. RF represents Random Forest; GBR represents Gradient Boosted Regression Tree}
\label{fig:rec}
\end{figure}

For regression problems, in order to calculate error analogously to how it is done in classification, we use a threshold ($\epsilon$) on the absolute error of primary predictions to be within tolerable limits. For example, as long as error is within $\pm \epsilon$ of the true value, it is considered an acceptable prediction (1). When the prediction of an algorithm is outside these bounds, it's considered an unacceptable (or incorrect) prediction (0). However, this threshold value is application specific and there is a need to detect a default value. To provide a default value, we use the null model concept introduced by \cite{rec}. Every regression problem has a null model and hence an REC curve associated with it. We detect the knee of this curve using the first convex dip in its double differential and choose the corresponding $\epsilon$ to be the default threshold value. An REC plot for the video dataset \cite{video} is shown in Figure \ref{fig:rec}. We calculate this default threshold for all the regression experiments reported in Section \ref{sec:expt}.

\section{Illustration}
\label{sec:expt}
We illustrate the performance of this algorithm using $5$ classification and regression datasets, listed in Table \ref{tab:classResutls} and \ref{tab:regResults} respectively. Features used by the MPP algorithm for the purpose of these experiments are same as the features used by the primary algorithm. Ideally, the score presented by {\it MPP} algorithm should match the predictive performance of the primary algorithm. It can be seen from the tables that the MPP algorithm is able to track the performance of primary algorithm in most of the datasets.

\begin{center}
    \begin{table}
    \centering
    \begin{tabular}{ | p{1.8cm} | p{1.5cm} | p{1.4cm} | p{1.4cm} | }
    \hline
    Dataset & Primary Algorithm Error & MPP predicted accuracy & Absolute difference\\ \hline
    Samsung \cite{Samsung} & 0.92 & 0.92 & 0.00\\ \hline
    Yelp \cite{yelp} & 0.95 & 0.95 & 0.00\\ \hline
    Census \cite{census} & 0.78 & 0.63 & 0.15\\ \hline
    Forest \cite{forest} & 0.65 & 0.64 & 0.01\\ \hline
    Letter \cite{letter} & 0.71 & 0.6 & 0.11\\ \hline
    \end{tabular}
    \caption{MPP's performance on classification datasets. Ideally, the primary algorithm accuracy and MPP's prediction should match.}
    \label{tab:classResutls}
    \end{table}
\end{center}

\begin{center}
    \begin{table}
    \centering
    \begin{tabular}{ | p{2.1cm} | p{1.5cm} | p{1.4cm} | p{1.1cm} |}
    \hline
    Dataset & Primary Algorithm Error & MPP predicted accuracy & Absolute difference\\ \hline
    Facebook \cite{facebook} & 0.56 & 0.56 & 0.00\\ \hline
    Songs \cite{songs} & 0.58 & 0.61 & 0.03\\ \hline
    Blog \cite{blog} & 0.73 & 0.71 & 0.02\\ \hline
    Turbine \cite{turbine} & 0.51 & 0.85 & 0.34\\ \hline
    Video \cite{video} & 0.59 & 0.72 & 0.13\\ \hline
    \end{tabular}
    \caption{MPP's performance on regression datasets with default epsilon value. Ideally, the primary algorithm accuracy (generated by thresholding with default epsilon) and the MPP's prediction should match}
    \label{tab:regResults}
    \vspace{-0.5cm}
    \end{table}
\end{center}

\vspace{-2cm}
\section{Conclusion}
We presented an approach {\it MPP} to track the predictive performance of a ML model in deployment. Such a score helps operations teams to create automated ML alerts and data scientists to get insights about the efficacy of deployed models in production. This helps both, the operations teams to monitor and manage the deployed ml model potentially preventing catastrophic predictions and the data scientists to get the information they need for further analysis of the production system. 

\newpage
\bibliographystyle{plain}
\bibliography{sample-bibliography}

\end{document}